\documentclass[conference]{IEEEtran}
\IEEEoverridecommandlockouts
\usepackage{cite}
\usepackage{amsmath,amssymb,amsfonts}
\usepackage{graphicx}
\usepackage{textcomp}
\usepackage{xcolor}
\def\BibTeX{{\rm B\kern-.05em{\sc i\kern-.025em b}\kern-.08em
    T\kern-.1667em\lower.7ex\hbox{E}\kern-.125emX}}
 
\usepackage{dblfloatfix}
\usepackage{cite}
\usepackage{amsmath,amssymb,amsfonts}
\usepackage{algpseudocode}
\usepackage{algorithm}
\usepackage{caption,subcaption}
\usepackage{hyperref}
\usepackage{float}
\usepackage{graphicx}
\usepackage{textcomp}
\usepackage{svg}
\usepackage{xcolor}
\usepackage{multirow}
\usepackage{array}
\usepackage{booktabs}
\usepackage{pifont}
\usepackage{subcaption}
\usepackage[export]{adjustbox}
\usepackage{multicol}
\usepackage{soul}

\usepackage{color} 
\usepackage{fancyhdr}
\newcolumntype{M}[1]{>{\centering\arraybackslash}m{#1}}

\title{\LARGE \bf
LRC-WeatherNet:  LiDAR, RADAR, and Camera Fusion  Network \\ for Real-time Weather-type Classification in Autonomous Driving}

\author{Nour Alhuda Albashir$^{1,\star}$, Lars Pernickel$^{1,\star}$, Danial Hamoud$^{1,\star}$, Idriss Gouigah$^{1}$,  and Eren Erdal Aksoy$^{1}$% <-this % stops a space
\thanks{*Co-funded by the European Union. Views and opinions expressed are however those of the author(s) only and do not necessarily reflect those of the European Union or European Climate, Infrastructure and Environment Executive Agency (CINEA). Neither the European Union nor the granting authority can be held responsible for them. Project grant no. 101069576.}% <-this % stops a space
\thanks{$^{1}$Halmstad University, School of Information Technology,
        Center for Applied Intelligent Systems Research, Halmstad, Sweden
        {\tt\small eren.aksoy@hh.se}}%
\thanks{$^\star$These authors contributed equally!}%
}
\begin{document}

\maketitle
\thispagestyle{empty}
\pagestyle{empty}

%%%%%%%%%%%%%%%%%%%%%%%%%%%%%%%%%%%%%%%%%%%%%%%%%%%%%%%%%%%%%%%%%%%%%%%%%%%%%%%%
\begin{abstract} 
Autonomous vehicles face major perception and navigation challenges in adverse weather such as rain, fog, and snow, which degrade the performance of LiDAR, RADAR, and RGB camera sensors. While each sensor type offers unique strengths—such as RADAR’s robustness in poor visibility and LiDAR’s precision in clear conditions—they also suffer distinct limitations when exposed to environmental obstructions. This study proposes LRC-WeatherNet, a novel multi-sensor fusion framework that integrates LiDAR, RADAR, and camera data for real-time classification of weather conditions. By employing both early fusion using a unified Bird's Eye View representation and mid-level gated fusion of modality-specific feature maps, our approach adapts to the varying reliability of each sensor under changing weather. Evaluated on the extensive MSU-4S dataset covering nine weather types, LRC-WeatherNet achieves superior classification performance and computational efficiency, significantly outperforming unimodal baselines in adverse conditions. This work is the first to combine all three modalities for robust, real-time weather classification in autonomous driving.
We release our trained models and source code in \url{https://github.com/nouralhudaalbashir/LRC-WeatherNet}. 
\end{abstract}

%%%%%%%%%%%%%%%%%%%%%%%%%%%%%%%%%%%%%%%%%%%%%%%%%%%%%%%%%%%%%%%%%%%%%%%%%%%%%%%%

%%%%%%%%%%%%%%%%%%%%%%%%%%%%%%%%%%%%%%%%%%%%%%%%%%%%%%%%%%%%%%%%%%%%%%%%%%%%%%%%

%%%%%%%%%%%%%%%%%%%%%%%%%%%%%%%%%%%%%%%%%%%%%%%%%%%%%%%%%%%%%%%%%%%%%%%%%%%%%%%%
\section{INTRODUCTION}
Autonomous vehicles (AVs) encounter substantial challenges in perception and navigation during adverse weather conditions such as rain, fog, and snow. 
These environmental factors significantly degrade the performance of key sensors, namely LiDAR, RADAR, and RGB camera, thus compromising the accuracy and reliability of downstream perception tasks such as scene classification and object detection. 

\begin{figure}[t!]
    \centering
    \includegraphics[width = 0.8\linewidth]{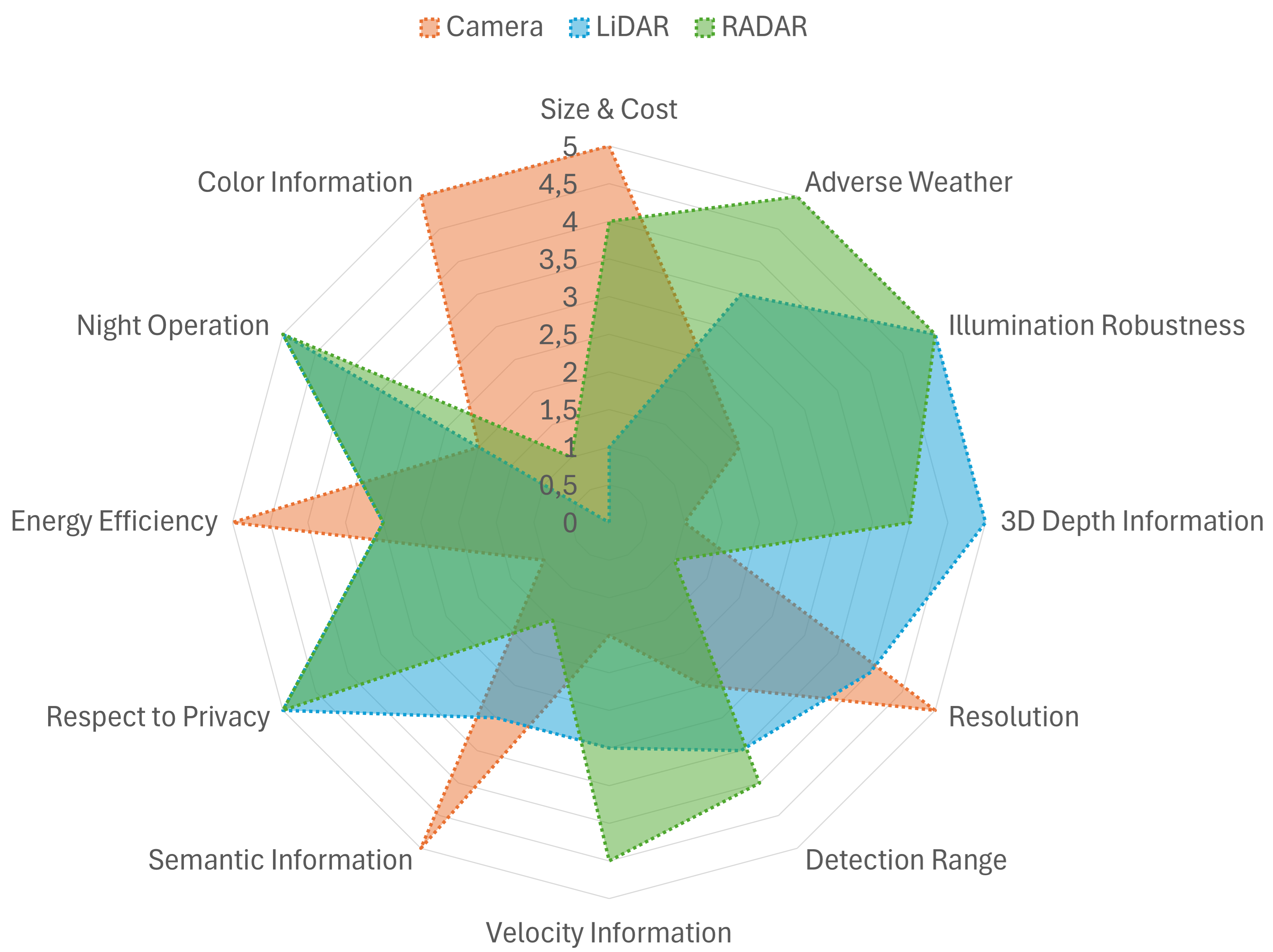}
    \caption{Illustration of the characteristics of LiDAR, RADAR, and Camera sensors. Contextually, LiDAR-RADAR-Camera fusion has the highest complementary properties.}
    \label{fig:sensor_char}
\end{figure}

In recent years, there has been growing interest in employing a diverse set of these sensor modalities for more reliable scene perception. 
Each sensor type, however, offers distinct advantages and limitations~\cite{18,2}, as illustrated in Fig.~\ref{fig:sensor_char}.
RADAR technology, for instance, maintains robust functionality in challenging weather such as heavy rain or fog, due to its reliance on long-wavelength electromagnetic signals that more effectively penetrate atmospheric obstructions  \cite{18}. 
While this resilience ensures stable operation, RADARs occasionally misinterpret reflections from nearby raindrops as false signals, introducing inaccuracies at close ranges. 
LiDARs, on the other hand, provide precise 3D perception under clear skies. Their effectiveness, however, diminishes significantly in dense fog or precipitation due to light scattering, particularly at longer detection ranges. 
Cameras excel at capturing rich visual cues, including color and contextual details, which are critical for interpreting scene semantics; however, their utility plummets in suboptimal weather conditions. 
Fusing readings from these complementary sensors and exploiting their unique information patterns reduces the shortcomings of each sensor, offering enhanced performance in challenging weather and seasonal conditions \cite{2}.

%-------------------------------------------------------------------------------
%-------------------------------------------------------------------------------

In this study, we address the challenge of robust weather condition classification for autonomous vehicles through a multi-sensor fusion framework. 
Leveraging the complementary sensing capabilities of LiDAR, RADAR, and camera, we propose a new model, named LRC-WeatherNet, that integrates readings from all three modalities in real-time to improve the accuracy of weather condition recognition. 
Our fusion pipeline explores both early fusion, where LiDAR and RADAR features are projected into a unified Bird's Eye View (BEV) representation, and mid-level gated fusion, which combines modality-specific feature maps after backbone processing.
Thus, our fusion strategy learns inherent strengths and weaknesses of each modality under different environmental conditions.
To assess model performance and sensor contributions, experiments are conducted using different backbone architectures (e.g., EfficientNet-B0~\cite{19} and PointPillars~\cite{lang2019pointpillars}).  

We evaluate the performance of LRC-WeatherNet on the Michigan State University Four Seasons (MSU-4S) dataset \cite{3}, which captures a wide range of seasonal and environmental conditions. The classification task includes nine distinct weather conditions, such as \textit{rain, spring snow, snow, fall, sunset, late summer, and early fall} scenarios.
We report a quantitative comparison of our model performance and computational efficiency using metrics such as accuracy, F1-score, confusion matrices, inference time, and the number of operations (measured in GMACs).
The obtained results show our lightweight BEV-based multi-modal fusion strategy significantly improves classification performance over different baselines, particularly under challenging weather conditions.

To the best of our knowledge, this is the first work to fuse LiDAR, RADAR, and camera data for real-time weather classification in autonomous driving. 
We release our trained models and source code \footnote{\url{https://github.com/nouralhudaalbashir/LRC-WeatherNet}}. 
%The source code and trained models will be made publicly available upon publication. 

%%%%%%%%%%%%%%%%%%%%%%%%%%%%%%%%%%%%%%%%%%%%%%%%%%%%%%%%%%%%%%%%%%%%%%%%%%%%%%%%%%%%%%%%

%%%%%%%%%%%%%%%%%%%%%%%%%%%%%%%%%%%%%%%%%%%%%%%%%%%%%%%%%%%%%%%%%%%%%%%%%%%%%%%%
\section{Related Work}
\label{sec:related}

A variety of fusion strategies have been developed in the literature. 
Early fusion, as in~\cite{11}, integrates the raw sensor readings prior to processing for the downstream perception task.
Early Fusion has the key advantage of not leaking any sensor readings because it relies exclusively on raw sensor data.
Late fusion, as in~\cite{6}, stands out as an alternative information aggregation method for merging high-level predictions derived individually from each modality. 
Contemporary research explores mid-level fusion strategies to capture descriptive scene features unique to each modality. 
Along these lines, the LiRa fusion in \cite{7} leverages LiDAR's dense spatial features and further incorporates RADAR's attributes to improve 3D object feature representation.
LiRa begins with the early fusion of LiDAR and RADAR voxel features extracted from intensity, velocity, and RADAR Cross-Section (RCS) data, which are jointly encoded using sparse convolutions. 
Next, in the middle fusion stage, LiRA introduces adaptive gated networks to assign channel-wise weights to the fused feature maps, enabling selective emphasis on sensor contributions based on contextual needs. 
Along the same lines, the work in~\cite{8} introduces a deep gated fusion approach to combine camera and LiDAR data for object detection.
These architectures demonstrate the potential of adaptive gated fusion networks to dynamically balance the contributions of sensors.
Our study extends this methodology by incorporating a camera with LiDAR and RADAR for weather classification in the 2D  Bird's Eye View (BEV) space.

DeepFusion in  \cite{9} begins by extracting high-level features from each sensor: LiDAR data is voxelized and converted into a BEV representation, while camera data is processed through CNNs to efficiently extract visual features for 3D object detection. 
These features are then fused and aligned at an advanced feature level, ensuring their complementarity. A key enhancement of DeepFusion is its learnable alignment layer, which adaptively aligns camera features with LiDAR voxels based on a cross-attention mechanism. 
%This mechanism assesses the relevance of each camera feature concerning each LiDAR voxel, ensuring the most relevant features from each modality are emphasized throughout the fusion process \cite{9}.

EarlyBird in~\cite{11}  proposes an early fusion framework for multi-view visual tracking, where image features from multiple cameras are projected into a shared BEV space and fused through a simple channel-wise concatenation prior to downstream processing. The architecture avoids explicit cross-attention or modality-specific alignment mechanisms, demonstrating that early spatial fusion can yield competitive performance while remaining computationally efficient.

Inspired by this paradigm, we extend EarlyBird's early fusion concept from multi-view visual inputs to multi-modal sensory inputs, more specifically LiDAR and RADAR. 
Unlike LiRa \cite{7}, in our approach, LiDAR and RADAR point clouds are independently projected into a common BEV space, and their modality-specific attributes (i.e., LiDAR intensity, RADAR SNR, and RADAR RCS) are stacked into a unified three-channel BEV representation. 
This design preserves spatial alignment across modalities while allowing the downstream convolutional backbone (EfficientNet~\cite{19}) to learn discriminative features from complementary sensory inputs. 
Following the work in \cite{7}, we introduce a gating mechanism to adaptively learn the early fused LiDAR and RADAR features in conjunction with those from the RGB camera sensor. 
By leveraging this architectural simplicity in the context of multi-modal data, we aim to balance representational richness with inference efficiency for weather-type classification tasks.

In the context of classifying diverse weather conditions, RECNet was introduced in~\cite{5} as a camera-only model for environmental condition classification.
Likewise, a lightweight network, RangeWeatherNet,  was proposed in~\cite{10} to classify both weather types and road conditions by using LiDAR-only data.
%
%For classification tasks, we refer to the work of Sebastian et al. \cite{10}, which investigates how varying weather conditions affect LiDAR point clouds by employing statistical data analyses to explore the impact of weather on LiDAR performance. The authors propose a novel lightweight network architecture, RangeWeatherNet, to address these challenges.
RangeWeatherNet processes LiDAR point clouds using a range image projection, converting 3D data into a 2D representation that is further processed through a modified DarkNet architecture~\cite{darknet}. 
Although this network enables efficient feature extraction while maintaining low computational overhead, it struggles with challenging scenarios such as fog and rain, often misclassifying dense fog as snow or light fog as clear weather \cite{10}. 
This limitation underscores the need for additional sensor data to improve robustness under adverse conditions.
Our study addresses these gaps by employing an adaptive data fusion framework that integrates LiDAR, RADAR, and camera modalities, thereby capitalizing on the complementary strengths of each sensor to enhance classification performance by prioritizing the most salient sensory information.

%%%%%%%%%%%%%%%%%%%%%%%%%%%%%%%%%%%%%%%%%%%%%%%%%%%%%%%%%%%%%%%%%%%%%%%%%%%%%%%%%%%%%%%%
\section{Method}

\begin{figure}[t!]
    \centering
    \includegraphics[width=\linewidth]{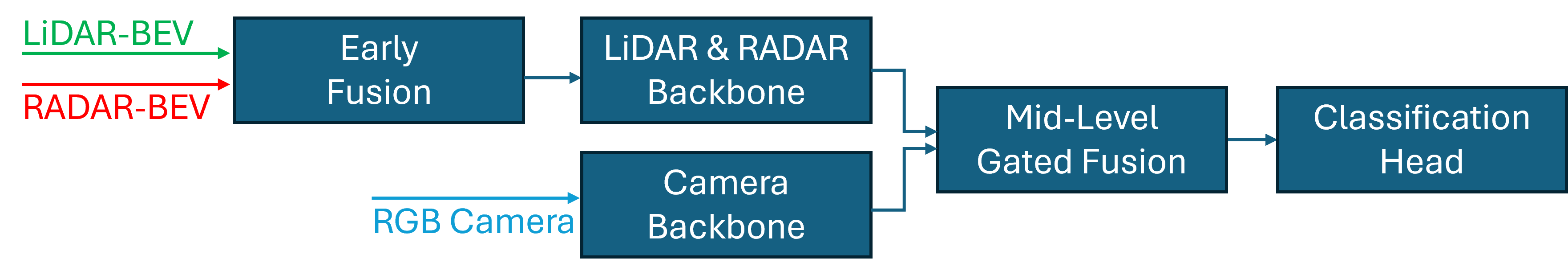}
    \caption{The LRC-WeatherNet architecture. Early fused LiDAR-RADAR BEV features are further combined with RGB features via a mid-level gated fusion module. Raw sensor outputs are passed through different backbone encoders.}
    \label{fig:architecture}
\end{figure}

%This section introduces how each sensor stream is processed independently in unimodal baseline experiments, and how fusion strategies, namely early fusion and mid-level gated fusion, are implemented to integrate information across modalities. 
Fig.~\ref{fig:architecture} illustrates the architectural design of our proposed multimodal fusion framework, LRC-WeatherNet.
The aim of our architectural design is to evaluate the effectiveness of the backbone models in modeling multimodal sensor input under varying environmental conditions, while maintaining computational efficiency and scalability.

Unlike previous approaches that rely on 3D point cloud backbones (PointPillars~\cite{lang2019pointpillars}, VoxelNeXt~\cite{chen2023voxelnext}, LiRa~\cite{7}), our framework utilizes EfficientNet-B0~\cite{19} to process 2D representations of multimodal sensor data.
By leveraging the high performance and efficiency of EfficientNet, the model can extract meaningful spatial and contextual features from BEV projections of LiDAR and RADAR point clouds, as well as from raw RGB images.

%-------------------------------------------------------------------------------
%-------------------------------------------------------------------------------

%%%%%%%%%%%%%%%%%%%%%%%%%%%%%%%%%%%%%%%%%%%%%%%%%%%%%%%%%%%%%%%%%%%%%%%%%%%%%%%%%%%%%%%%
\subsection{Data Representation}
Raw LiDAR and RADAR point clouds are converted into 2D BEV representations. This transformation simplifies the inherently sparse and unstructured 3D data into a dense, top-down grid format, facilitating the extraction of spatially coherent patterns by convolutional neural network backbones. 

LiDAR and RADAR point clouds are confined to a predefined frontal region aligned with the vehicle's forward field of view to minimize noise and computational complexity. %The following parameters govern the BEV projection for both modalities:
In the LiDAR frustum filtering,  the longitudinal range ($x$) is set to $[0,50]$$~\text{m}$ (forward direction), while the lateral range ($y$) is $[-25,25]$$~\text{m}$ (sideways). 
This $50 \times 50\,\mathrm{m}^2$ frustum approximates the vehicle's immediate operational zone, excluding extraneous data from rear and peripheral regions. 
Next, each 3D LiDAR point is mapped to a 2D cell based on a fixed grid resolution defined as $x_{\text{bev}} = \left\lfloor \frac{x - x_{\text{min}}}{\text{resolution}} \right\rfloor, \quad 
y_{\text{bev}} = \left\lfloor \frac{y - y_{\text{min}}}{\text{resolution}} \right\rfloor$, with coordinates clipped at the boundary limits.
A grid resolution of $0.1\,\mathrm{m}$ is used to ensure sufficient granularity in BEV, resolving delicate environmental structures while minimizing computational overhead. 
In LiDAR BEV, the maximum intensity value is taken per cell to emphasize dominant surface returns. 
Finally, the raw $500 \times 500$ BEV grid is downsampled to $224 \times 224$ pixels via bilinear interpolation, aligning with the input specifications of the backbone architecture.

RADAR data, stored as 5D vectors $(x, y, z, \text{SNR}, \text{RCS})$,  undergo a similar preprocessing stage with modality-specific adaptations. 
The region filtering is applied to retain forward-facing returns within the bounds $x = [0,50]~\text{m}$ and $y = [-25,25]~\text{m}$.
Next, two distinct BEV maps are constructed: SNR and RCS maps. 
The SNR map encodes signal-to-noise ratios to highlight reliable detections.
The RCS map is used for storing RADAR cross-section values that correlate with material reflectivity and moisture content.
After retaining the maximum values per cell to emphasize dominant RADAR signatures, both SNR and RCS maps are combined into a $[2, H, W]$ tensor and resized to $224 \times 224$ pixels.

%%%%%%%%%%%%%%%%%%%%%%%%%%%%%%%%%%%%%%%%%%%%%%%%%%%%%%%%%%%%%%%%%%%%%%%%%%%%%%%%%%%%%%%%
\subsection{Early Fusion of LiDAR and RADAR }

LiDAR and RADAR BEV representations are fused at the input level using timestamp-based alignment and channel-wise concatenation to exploit complementary sensor signatures. This approach ensures temporal and spatial coherence between modalities while retaining raw sensor-specific features critical for detecting weather-induced distortions.

Formally, for a LiDAR tensor $L \in \mathbb{R}^{1 \times H \times W}$ and a RADAR tensor $R \in \mathbb{R}^{2 \times H \times W}$, the fused tensor $F$ is defined as $F = \text{Concat}(L, R) \in \mathbb{R}^{3 \times H \times W}$,
where $H = W = 224$ aligns with the input requirements of the backbone model. 
This early fusion strategy allowed the backbone network to jointly learn cross-modal feature representations from LiDAR and RADAR within a shared input space.

%%%%%%%%%%%%%%%%%%%%%%%%%%%%%%%%%%%%%%%%%%%%%%%%%%%%%%%%%%%%%%%%%%%%%%%%%%%%%%%%%%%%%%%%
\subsection{Backbone Models}
%The BEV representation of the early fused LiDAR and RADAR streams is passed to the backbone feature extractor, which is a customized EfficientNet-B0 model~\cite{19}, \red{pretrained on ImageNet}. 

There exist two backbone models: one is for the early fused LiDAR and RADAR data, while the other is for the raw RGB camera images (see Fig.~\ref{fig:architecture}).
Both backbones are based on the EfficientNet-B0 model~\cite{19}, initialized with pre-trained ImageNet~\cite{deng2009imagenet}  weights without altering the architecture. 
%An early fusion backbone is constructed to evaluate the benefits of combining complementary sensor modalities by integrating LiDAR and RADAR BEV features at the input level. 
%The fused tensors contained three channels: LiDAR intensity, RADAR SNR, and RADAR RCS, stacked along the channel dimension to form a unified input of shape $[3 \times 224 \times 224]$. 
%Since this format matches the original RGB input configuration of EfficientNet-B0, no modification to the model's first convolutional layer was necessary. 
%The model was initialized with pretrained ImageNet weights without altering its architecture.
%
%The backbone for the camera image stream is the same EfficientNet-B0 model~\cite{19}, pretrained on ImageNet~\cite{deng2009imagenet}. 
All RGB inputs are resized to $224 \times 224$ and normalized using the ImageNet statistics.

%%%%%%%%%%%%%%%%%%%%%%%%%%%%%%%%%%%%%%%%%%%%%%%%%%%%%%%%%%%%%%%%%%%%%%%%%%%%%%%%%%%%%%%%
\subsection{Mid-Level Gated Fusion}

The effective integration of spatial information derived from camera images with early fused LiDAR and RADAR sensor data is crucial, yet it poses significant challenges due to the unique  characteristics and representations of these sensor data.
To address this, our proposed LRC-WeatherNet architecture, as shown in Fig.~\ref{fig:architecture}, employs a middle fusion strategy at the intermediate feature level, after backbone feature extraction but before the final classification. 

Inspired by works in~\cite{7,Yoo_2020,Kim2018}, the middle fusion mechanism in our approach employs a gated fusion module, which dynamically modulates the contribution of each modality to the final fused representation. Instead of a simple feature concatenation, the gated fusion mechanism learns spatially and channel-wise adaptive weights during training. These weights are computed from the joint feature representations extracted by the backbone modules, thereby enabling the model to assess the relative importance of the early fused LiDAR and RADAR as well as camera features. 
This adaptive weighting enables the network to prioritize the most informative modality while suppressing the influence of those that provide less relevant or noisier signals.

%-------------------------------------------------------------------------------
\begin{figure}[b!]
    \centering
    \includegraphics[width=\linewidth]{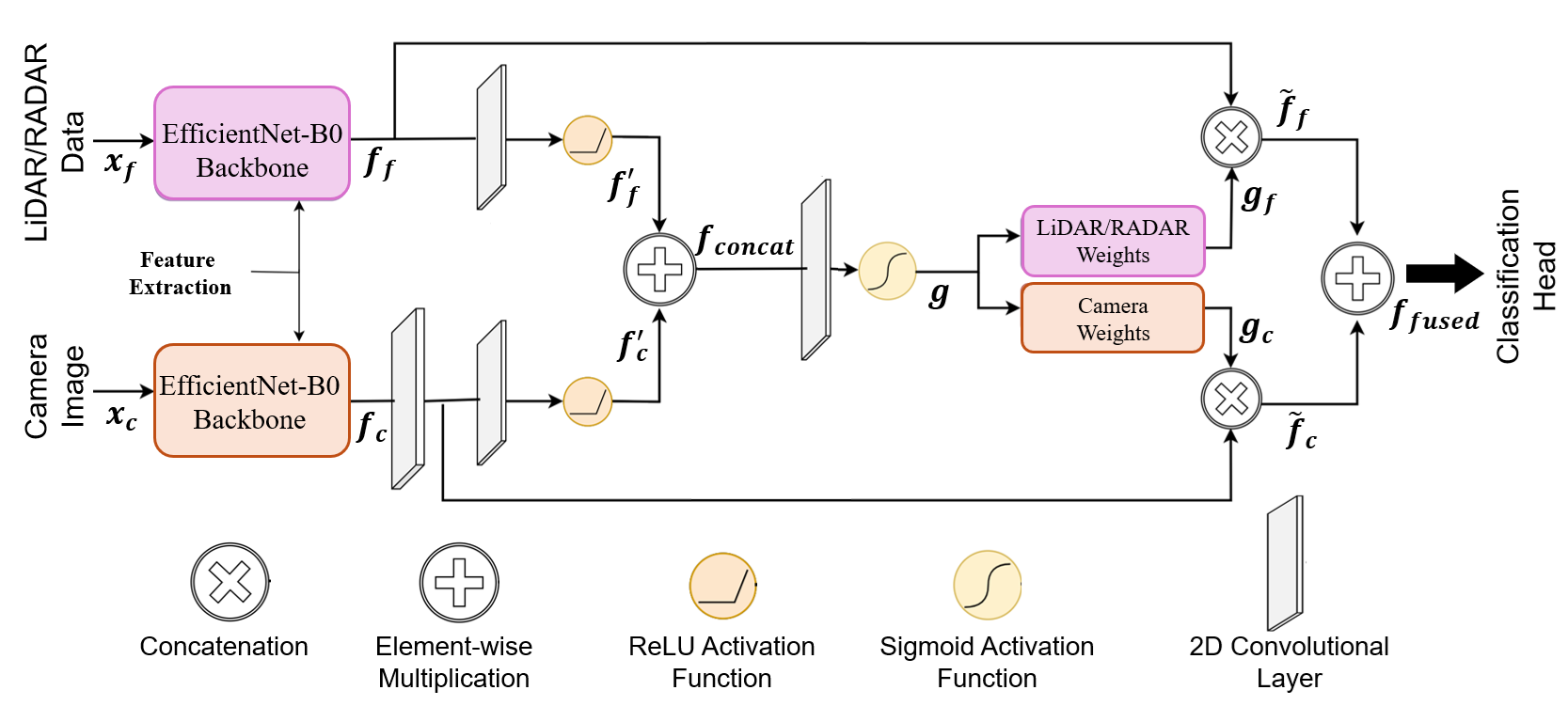}  % adjust file path
    \caption{Illustration of the gated fusion mechanism, where adaptive weights are computed to control the contributions of each modality. }
    \label{fig:gating_module}
\end{figure}
%-------------------------------------------------------------------------------

The detailed architecture of our gated fusion mechanism is illustrated in Fig.~\ref{fig:gating_module}. 
More specifically, let the inputs be $\mathbf{x}_f \in \mathbb{R}^{3 \times H \times W}$ and  $\mathbf{x}_c \in \mathbb{R}^{3 \times H \times W}$ for the fused LiDAR and RADAR BEV data ($\mathbf{x}_f$) and camera streams ($\mathbf{x}_c$), respectively.
Each input is passed through a dedicated EfficientNet-B0 backbone to compute the feature vectors  as  $\mathbf{f}_f = \text{EffNet}_f(\mathbf{x}_f) \in \mathbb{R}^{d}$ and $\mathbf{f}_c = \text{EffNet}_c(\mathbf{x}_c) \in \mathbb{R}^{d}$, where $d = 1280$ is the feature dimension after global average pooling.
Next, each modality feature vector is first projected through its own MLP as $ \mathbf{f}_f' = \text{ReLU}(W_f \mathbf{f}_f + \mathbf{b}_f)$ and $\mathbf{f}_c' = \text{ReLU}(W_c \mathbf{f}_c + \mathbf{b}_c)$.

These projected features are then concatenated as $\mathbf{f}_{\text{concat}} = [\mathbf{f}_f', \mathbf{f}_c'] \in \mathbb{R}^{2d}$.
This vector is passed through a gating MLP as $g = \sigma(W_g \mathbf{f}_{\text{concat}} + \mathbf{b}_g) \in \mathbb{R}^{2d}$.

The gating vector is then split back to individual components $g = [g_f, g_c], \quad g_f, g_c \in \mathbb{R}^{d}$. Each of these components is then gated with the original feature vectors as 
    $\tilde{\mathbf{f}}_f = \mathbf{f}_f \odot g_f$ and  $\tilde{\mathbf{f}}_c = \mathbf{f}_c \odot g_c$.

Finally, both gated features are concatenated $\mathbf{f}_{\text{fused}} = [\tilde{\mathbf{f}}_f, \tilde{\mathbf{f}}_c] \in \mathbb{R}^{2d}$ to be passed through the classification head.

The gating module employs a sigmoid-activated linear layer to compute modality-specific weighting coefficients. These weights are applied element-wise to the original (non-projected) feature vectors, ensuring preservation of input fidelity. The adaptively weighted features are concatenated and propagated through a unified two-layer classifier to generate predictions across different weather classes.

%%%%%%%%%%%%%%%%%%%%%%%%%%%%%%%%%%%%%%%%%%%%%%%%%%%%%%%%%%%%%%%%%%%%%%%%%%%%%%%%%%%%%%%%
\subsection{The Classification Head}

To ensure a fair comparison with different baseline unimodal and multimodal experiments, the original classification head of both backbones was replaced with a custom classifier comprising:
\begin{itemize}
    \item a 512-unit dense layer with Batch Normalization and ReLU activation,
    \item a dropout layer (dropout rate = 0.3), and
    \item a 9-unit output layer for softmax-based multi-class classification.
\end{itemize}

More specifically, the final fused feature vector ($\mathbf{f}_{\text{fused}}$) is passed through a two-layer MLP as $\hat{\mathbf{y}} = W_2 \cdot \text{ReLU}(\text{BN}(W_1 \cdot \mathbf{f}_{\text{fused}} + \mathbf{b}_1)) + \mathbf{b}_2$
Here, $\hat{\mathbf{y}} \in \mathbb{R}^{9}$ denotes the predicted class logits. $\text{BN}$ represents Batch Normalization. Dropout is applied before the final layer.

%%%%%%%%%%%%%%%%%%%%%%%%%%%%%%%%%%%%%%%%%%%%%%%%%%%%%%%%%%%%%%%%%%%%%%%%%%%%%%%%%%%%%%%%
\section{Experiments}

%%%%%%%%%%%%%%%%%%%%%%%%%%%%%%%%%%%%%%%%%%%%%%%%%%%%%%%%%%%%%%%%%%%%%%%%%%%%%%%%%%%%%%%%
\subsection{Dataset}
We used the Michigan State University Four Seasons (MSU-4S) dataset introduced in \cite{3}, including 100K data samples from various challenging weather scenarios, such as rain and snow.
The dataset was recorded over different time periods in four seasons of the year across five distinct geographic regions: thoroughfare (light blue), forest (green), industrial (red), central hub (dark blue), and neighborhood (yellow). 
Each region reflects a unique environmental context. 
Most of the data was collected during the evening hours, but some sessions were recorded at other times of day.

Dataset splits were stratified into training, validation, and test sets across nine weather classes. The total number of samples in each split is as follows: Train set: 96{,}567 files (60\%), Validation set: 32{,}188 files (20\%), and Test set: 32{,}197 files (20\%).
During the partitioning process, instead of applying a naive random splitting, we ensured that samples from the same driving sessions remained within the same partition to maintain the temporal structure of the data.

The dataset comprises two LiDAR sensors, six RADAR sensors (Continental ARS430), three cameras, and supplementary odometry sensors. 
For simplicity, we adopted a more constrained evaluation setup, utilizing only the front-view streams from the RADAR and LiDAR (Ouster OS-1, 64-line) sensors together with a single front-facing camera.
%The LiDAR sensor is the Ouster OS-1 64-line 3D LiDAR, which is roof-mounted near the front-left and offers a 360-degree field of view for capturing detailed 3D point clouds of the surrounding space. 
%In addition, three front cameras have been installed in the vehicle, covering a combined 150-degree field of view. 
%
The dataset was divided into nine classes, each representing a specific combination of season and weather condition. 
These classes capture both inter-seasonal differences and intra-seasonal variations.

%The MSU-4S dataset includes six Continental ARS430 medium-to-long-range RADARs strategically mounted around the vehicle to achieve full 360-degree coverage. These RADARs provide individual object detections along with RADAR Cross Section (RCS), Signal-to-Noise Ratio (SNR), and coordinate information.

Sensor data lacking corresponding measurements from other sensors at the same timestamp were excluded to ensure consistency and reliability in weather classification. This filtering step guarantees that only fully synchronized data samples are used in multi-sensor fusion experiments.

\subsection{Training Protocol}

The model was trained using the AdamW optimizer with a learning rate of $3 \times 10^{-4}$ and weight decay of $1 \times 10^{-4}$. Learning rate scheduling was applied to reduce the learning rate when validation loss plateaued.

To mitigate class imbalance, class-weighted cross-entropy loss was employed~\cite{cortinhal2020salsanext}. Class weights were computed inversely proportional to the class frequency in the training set and clamped to avoid extreme values. Gradient clipping (max norm = 5.0) was applied to stabilize the training process.

All experiments were conducted on an Ubuntu 24.04.2 LTS system with an AMD Ryzen Threadripper PRO 7945WX 12-Core CPU, an NVIDIA RTX 4000 Ada Generation GPU (20 GB VRAM, driver version 575.57.08, CUDA 12.9), and 64 GB of RAM.

\subsection{Baselines} 
%In addition to the recent weather-type classification models such as RECNet~\cite{5} and RangeWeatherNet~\cite{10},  we trained various unimodal and multimodal baselines.
In addition to the recent weather detection model RECNet~\cite{5},  we trained several uni- and multimodal baselines.

%--------------------------------------------------------------------------
\begin{table*}[!b]
\caption{Quantitative performance evaluation across different unimodal and fusion configurations.}
\label{tab:efficientnet_results}
%\normalsize
\begin{center}
\resizebox{0.8\textwidth}{!}{
\begin{tabular}{lcccccc}
\toprule
Model & Val. Acc. & Test Acc. & Macro F1 & Inference (ms) & Params & GMAC \\
\midrule
Camera-only & 91.96\% & 77.86\% & 0.76 & 4.86 & 4.67M & 0.41 \\
LiDAR-only & 57.46\% & 56.34\% & 0.55 & 3.52 & 4.67M & 0.41 \\
RADAR-only & 31.99\% & 27.32\% & 0.26 & 3.79 & 4.67M & 0.41 \\
Early Fusion (LiDAR + RADAR) & 62.98\% & 61.06\% & 0.59 & 4.21 & 4.67M & 0.41 \\
%+ Camera Concatenation & 94.01\% & 86.52\% & 0.85 & 6.93 & 9.33M & 0.83 \\
%+ Camera Mid-Fusion & 93.82\% & 86.66\% & 0.85 & 7.13 & 19.17M & 0.84 \\
LRC-WeatherNet (Ours)  & 93.82\% & 86.66\% & 0.85 & 7.13 & 19.17M & 0.84 \\
\midrule
{RECNet~\cite{5}} & 20.40\% & 30.91\% & 0.22 & 74.6 & 77.8 M & 1.36 \\
%\red{RangeWeatherNet~\cite{10}} & 00.00\% & 00.00\% & 0.00 & 0.00 & 0.00 M & 00.00 \\

%LiDAR-only-PP & 67.99\% & 74.76\% & 74.76 & 8.57 & 4.86 M & 29.57 \\
%RADAR-only-PP & 36.31\% & 31.91\% & 0.3142 & 3.51 & 4.86 M & 01.85 \\
%Early Fusion-PP (LiDAR + RADAR) & 97.29\% & 46,33\% & 42,99 & 8.58 & 4.86 M & 29.57 \\
%LRC-WeatherNet-PP & 91.67\% & 87.77\% & 0.87 & 64.40 & 8.68 M & 88.67 \\
LiDAR-only-PP & 83.11\% & 81.49\% & 0.81 & 8.57 & 4.86 M & 29.57 \\
RADAR-only-PP & 43.80\% & 37.80\% & 0.37 & 3.51 & 4.86 M & 01.85 \\
Early Fusion-PP (LiDAR + RADAR) & 82.10\% & 82.30\% & 0.81 & 8.58 & 4.86 M & 29.57 \\
LRC-WeatherNet-PP & 91.67\% & 87.77\% & 0.87 & 64.40 & 8.68 M & 88.67 \\
\bottomrule
\end{tabular}
}
\end{center}
\end{table*}
%--------------------------------------------------------------------------

\textbf{Unimodal:} 
We trained LiDAR-only, RADAR-only, and Camera-only models using the same EfficientNet-B0 backbone for a fair and consistent comparison.  %The classifier head, pretrained weights, optimizer settings, loss function, and evaluation protocol were identical for a controlled comparison.
The only architectural difference lies in the input configuration: the first convolutional layer of EfficientNet-B0, originally defined for 3-channel RGB inputs, was modified to accept single-channel input, corresponding to the LiDAR intensity-only BEV representation in the LiDAR-only modal. Specifically, the original 3-channel kernel weights were averaged across the input dimension to initialize a new \texttt{Conv2d} layer with a single input channel. This preserved compatibility with the pretrained ImageNet weights while allowing the model to accept LiDAR tensors of shape $[1 \times 224 \times 224]$. The input data was normalized using dataset-specific statistics: mean = 0.0471, std = 0.1659.
In the RADAR-only model, the first convolutional layer of EfficientNet-B0 was modified to accept 2-channel RADAR BEV tensors, replacing the default 3-channel RGB input. The pretrained weights for this layer were adapted by averaging the original weights across the RGB channels.
The input data was normalized using precomputed dataset-specific statistics: mean = $[0.0072, 0.0040]$, std = $[0.2507,0.2326]$.
All other model components,  optimizer settings, loss function, and evaluation protocol remained unchanged, ensuring that performance differences arise solely from the input modality, rather than variations in architecture or training procedure.

During training, extensive data augmentation was applied to the camera stream to improve generalization. The augmentation includes horizontal and vertical flips, random rotations ($\pm 45^\circ$), color jittering, affine translations, and random resized cropping.
This augmentation was applied to both the Camera-only and multimodal fusion pipelines to ensure fairness in input variability.

Given the sparse and noisy nature of RADAR data, a moderate augmentation pipeline, including random affine rotations ($\pm 5^\circ$) and  Gaussian noise injection (2\% noise intensity), was applied during training. 
These augmentations aimed to simulate variability in RADAR returns caused by sensor drift and minor environmental changes, thereby enhancing the model's generalization ability.

\textbf{LRC-WeatherNet-PP:}
We further evaluated the performance of our proposed LRC-WeatherNet using a PointPillars-based backbone for the early fused LiDAR and RADAR readings, while preserving the camera branch. 

This architecture, named LRC-WeatherNet-PP, processes raw 3D point clouds into pillar-encoded pseudo-images, allowing the model to retain multi-scale geometric detail through spatially aware feature maps at different resolutions.
PointPillars~\cite{lang2019pointpillars} represents point cloud data using vertical columns, referred to as \emph{pillars}, on the x--y plane. Unlike voxel-based models that discretize all three spatial dimensions, PointPillars preserves the z-dimension (height) as a continuous feature within each point. 
%While the final BEV representation is 2D, the z-values are encoded as part of the input feature vector, allowing the network to model local height variations and 3D geometry within each pillar. 
Each point is described by nine input features: the four original attributes (x, y, z, intensity), three offsets from the mean point within the pillar ($\Delta$x, $\Delta$y, $\Delta$z), and two offsets from the pillar center in the x--y plane. These features are passed through a learnable encoding layer that transforms raw point features into richer point-wise representations, which are then aggregated and processed using efficient 2D convolutions. This design allows the model to effectively capture 3D structure while maintaining computational efficiency during inference.
Unlike PointPillars, the BEV transformation completely removes the z-dimension, resulting in purely 2D inputs. 
%These inputs encode sensor-derived features, such as intensity, SNR, and RCS, projected onto the x--y grid. This simplification reduces input complexity but also sacrifices vertical structural information.

\subsection{Quantitative Results} \label{sec:efficientnet_results}

A summary of the quantitative results on MSU-4S \cite{3} is provided in Table~\ref{tab:efficientnet_results}. 
%Each sensor model was evaluated on a held-out test set composed of nine distinct weather classes. 
Metrics reported include test accuracy, macro-averaged F1 score, inference time, number of parameters, and computational cost (as measured by GMAC).

The RADAR-only baseline model exhibited limited classification capability when using RADAR BEV inputs exclusively. It achieved a test accuracy of 27.32\% and a macro-averaged F1 score of 0.26. 
This low performance reflects the limited discriminative power of RADAR SNR and RCS maps, which encode low-dimensional motion and reflectivity patterns but lack fine spatial structure or texture. Furthermore, RADAR suffers from low vertical resolution and multipath interference, particularly in urban environments, where reflections from the ground or vehicles may mimic precipitation effects.

%RADAR is relatively robust to weather interference, unlike LiDAR, but that robustness does not translate into discriminability across visually or seasonally complex scenes. RADAR's inability to resolve geometry means that environmental cues like leaf density, snow cover, or lighting changes offer little benefit, resulting in high intra-class confusion.

The LiDAR-only baseline demonstrated a substantial improvement over the RADAR-only approach, achieving a test accuracy of 56.34\% and a macro-averaged F1 score of 0.55. 
This limited performance stems from scene-level clutter or reduced LiDAR returns resulting from reflection loss, particularly in snowy conditions. 
Snow essentially introduces scattering artifacts that reduce point cloud density, leading to mispredictions into rain or spring snow categories.

The Camera-only model achieved the strongest results among all unimodal baselines, with a test accuracy of 77.86\% and a macro-averaged F1 of 0.76.
RGB excelled in scenarios where chromatic and semantic richness were present, e.g., vibrant vegetation. The model leveraged textural differences and lighting variations to distinguish between otherwise geometrically similar scenes.
While this baseline excelled under well-lit, high-contrast conditions, its performance degraded in settings affected by glare or low ambient light.

%In summary, RADAR is robust to weather but lacks spatial expressiveness. LiDAR provides rich geometric data but struggles in occluded, snowy, or visually ambiguous urban scenes. The camera sensor achieved the highest performance thanks to rich semantics, but is vulnerable to lighting instability.
%These results justify the pursuit of multimodal fusion strategies, where LiDAR provides structure, camera provides semantics, and RADAR adds robustness. 

Combining LiDAR and RADAR BEV features through our early fusion improved performance relative to RADAR-only and LiDAR-only unimodal baselines, reaching a test accuracy of 61.06\% and a macro F1 score of 0.59. 
Our early fusion model also maintains computational efficiency, with 4.67 million parameters and an inference time of just 4.21~ms, while providing improved discriminability in challenging conditions.
%The fusion mechanism leverages RADAR's robustness in precipitation and LiDAR's structural precision in clear weather. This synergy is particularly evident in classes that performed poorly in unimodal configurations.

%Concatenating RGB image features with early-fused LiDAR and RADAR representations yielded 86.5\% test accuracy and a macro-averaged F1 score of 0.8552.
Finally, in our proposed LRC-WeatherNet, the mid-level gated fusion, combining early fused LiDAR and RADAR features with camera image features, produced the best overall performance, achieving 86.66\% test accuracy and a macro-averaged F1 score of 0.85.

Our baseline experiments (LRC-WeatherNet-PP), where a PointPillars-based backbone is used for the early fused LiDAR and RADAR data, achieved a classification accuracy of 87.77\% and an F1 score of 0.87, demonstrating the effective integration of spatial and visual modalities. However, the LRC-WeatherNet-PP baseline becomes substantially expensive due to extremely high computation time (64.40 ms) in contrast to 7.13 ms for the original LRC-WeatherNet, as shown in Table~\ref{tab:efficientnet_results}.

%\red{DANIAL: In your LiDAR-only and RADAR-only models, you used frustum filtering for 80 meters. However, Nour used 50 meters for x and y...can you retrain your models with the same filtering parameters?  $x = [0,50]~\text{m}$ and $y = [-25,25]~\text{m}$. }
%
%\red{DANIAL: How come the LRC-WeatherNet-PP model has 64.4 ms....the early fusion one takes 8.58...lets retrain  LRC-WeatherNet-PP by using EfficientNet-B0 instead of EfficientNet-B7...}
%
%\red{We should train the RangeWeatherNet model in~\cite{10}  and the RECNet model in~\cite{5} on MSU-4S and add the results to Table~\ref{tab:efficientnet_results}.}

%\red{The RECNet model in~\cite{5} achieved XXXX  on MSU-4S as shown in Table~\ref{tab:efficientnet_results}.}
Note that the camera-only model RECNet~\cite{5} achieved relatively low accuracy and high inference time, even compared to our camera-only baseline. This shows the strong performance of the camera branch of LRC-WeatherNet.  

%--------------------------------------------------------------------------
\begin{figure}[b!]
    \centering
    \includegraphics[width=\linewidth,height=9cm,keepaspectratio]{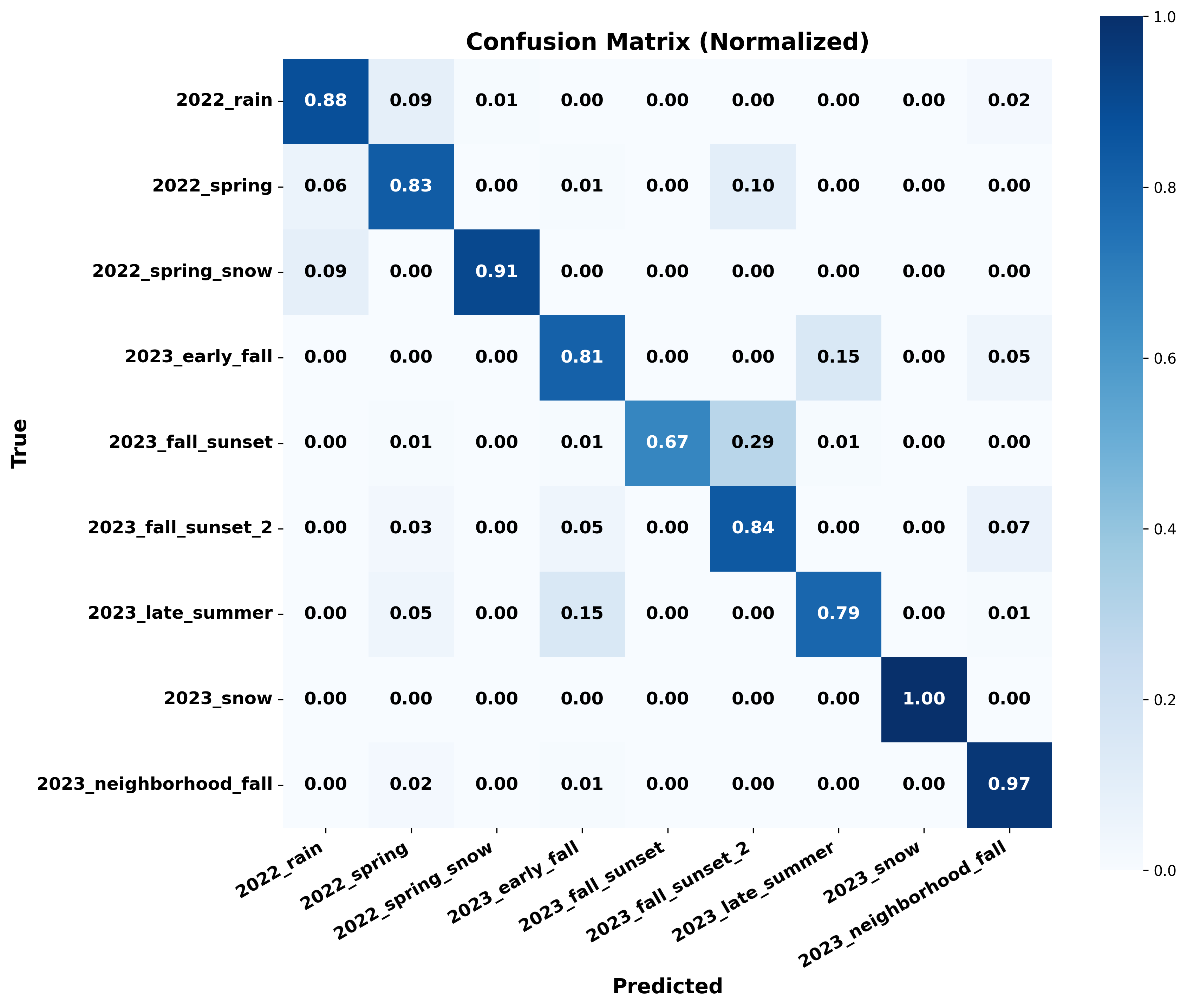}
    \caption{Confusion matrix of the proposed LRC-WeatherNet model predictions on MSU-4S.}
    \label{fig:confmat_bestfusion}
\end{figure}
%--------------------------------------------------------------------------

Fig.~\ref{fig:confmat_bestfusion} illustrates the prediction distribution of the proposed  LRC-WeatherNet fusion model across all weather classes in a confusion matrix.
As shown in this confusion matrix, our model LRC-WeatherNet exhibits high diagonal dominance and low off-diagonal noise.
Class-wise predictions are notably sharp, reflecting stronger model confidence and fewer ambiguous misclassifications. 
The model's confusion matrix indicates clearer separation between visually or seasonally adjacent categories, with low cross-class confusion.
These gains are largely attributed to the gated fusion strategy, which enables the network to dynamically control the contribution of each modality. 
The gated mechanism learns context-dependent weights, emphasizing features that are more informative for a given scene while attenuating noise from less relevant channels. 
This allows the network to preserve LiDAR's structural precision, RADAR's weather robustness, and RGB camera's semantic richness in a coherent, task-aware representation.

\subsection{Qualitative Resutls}

\begin{figure}[b!]
    \centering
    \includegraphics[width = \linewidth]{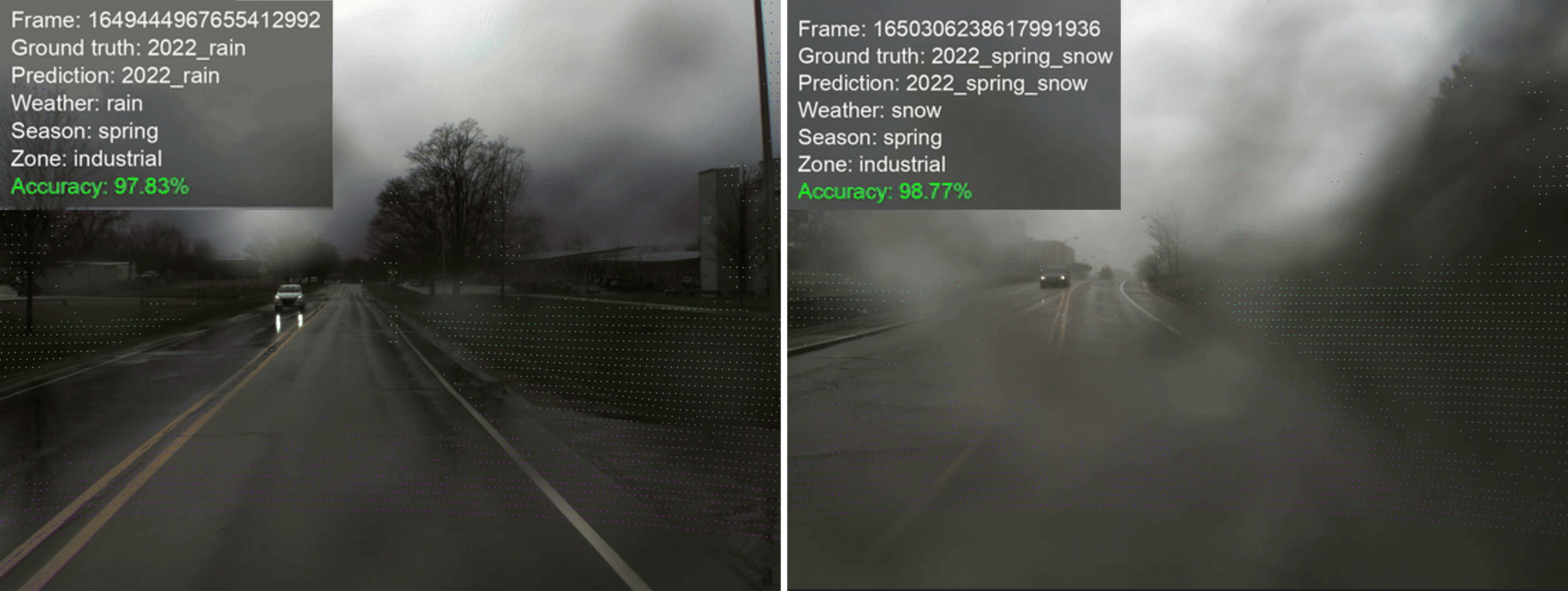}
    \caption{A sample qualitative result showing a spring rainy (left) and spring snowy (right) scenes predicted accurately.}
    \label{fig:qualitative_results}
\end{figure}

Fig.~\ref{fig:qualitative_results} depicts two sample images from the  MSU-4S dataset \cite{3} for the qualitative evaluation. 
Here, early fused LiDAR and RADAR point clouds are superimposed on the RGB camera image, and the proposed LRC-WeatherNet correctly predicts the right weather type for each scene. 
%\red{In the supplementary video, we provide more qualitative results}

%--------------------------------------------------------------------------
%--------------------------------------------------------------------------

\begin{figure}[t!]
    \centering
    \includegraphics[width = \linewidth]{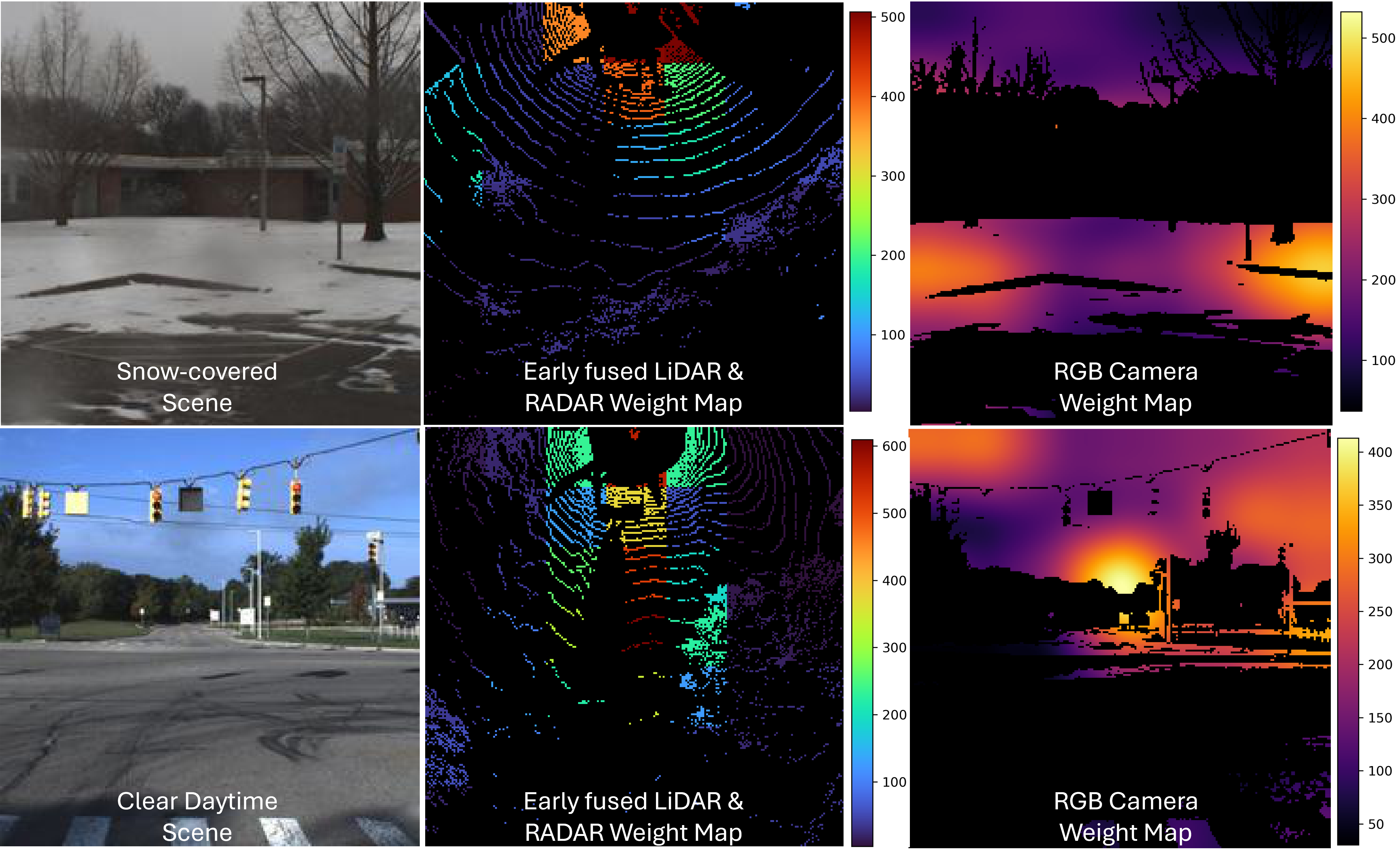}
    \caption{Visualization of weight maps learned in the gated fusion mechanism for two sample scenes: snowy (top row) and clear daytime (bottom row) scenes.  }
    \label{fig:gated_visual}
\end{figure}

To elucidate the internal dynamics of LRC-WeatherNet, we analyzed the modality-specific contributions facilitated particularly by the mid-level gated fusion mechanism. Fig.~\ref{fig:gated_visual} illustrates how the model adeptly balances spatial information from the early fused LiDAR and RADAR data with semantic content from RGB camera imagery, adapting to the contextual demands of each scene.
The top-left image in Fig.~\ref{fig:gated_visual} presents the RGB camera image of a snow-covered scene, characterized by flat lighting and low-contrast textures. 
In this specific instance of a snowy scene, the model accurately predicted the right class with the highest probability. The gated fusion mechanism assigned weights of $0.64$ to RGB and $0.35$ to the early fused LiDAR and RADAR, indicating a higher reliance on visual cues in this snowy environment.
The corresponding RGB weight map (Fig.~\ref{fig:gated_visual} top-right) reveals heightened activation in the lower regions of the image, particularly where snow accumulation and surface reflections are prominent. This suggests that the model leverages visual patterns, such as brightness and texture variations, to identify snow-covered areas.
Conversely, the early fused LiDAR and RADAR weight map (Fig.~\ref{fig:gated_visual} top-middle) exhibits strong activation along dense point cloud returns, especially near the foreground and building contours. This reflects the structural information captured by the fused spatial modalities, which complements the visual data, particularly in areas where visual cues may be less distinct.

%--------------------------------------------------------------------------
%--------------------------------------------------------------------------

The bottom-left image in Fig.~\ref{fig:gated_visual} shows a clear daytime scene at an intersection, with the model predicting the right class with the highest probability. The gated fusion weights were $0.43$ for RGB and $0.56$ for the early fused LiDAR and RADAR data, indicating a stronger reliance on the LiDAR and RADAR features.
This clear daytime scene image displays moderate visual cues such as green trees and a clear sky, but lacks prominent seasonal indicators. In contrast, the early fused LiDAR and RADAR contribution map in Fig.~\ref{fig:gated_visual} bottom-middle reveals dense and layered return patterns, characteristic of tree canopies and structural road elements. These geometric and reflectivity-based patterns provide highly distinctive signatures that enable the model to recognize early fall scenes.
This illustrates how, in environments with limited semantic variability in RGB images, the model shifts attention toward spatial and reflective structure captured in 3D BEV data. The LiDAR and RADAR stream provided more discriminative information in this case, allowing the model to confidently identify the correct seasonal class.
Taken together, these examples highlight the adaptive nature of the gated fusion mechanism: it does not apply fixed weights to each modality, but instead responds to scene-specific characteristics by emphasizing the sensor stream offering the most informative features. This dynamic allocation of attention enables the model to operate effectively across diverse visual and structural conditions and is a key factor behind its high classification accuracy and minimized inter-class confusion.

These learned weight map examples in Fig.~\ref{fig:gated_visual} highlight the adaptive behavior of the proposed  LRC-WeatherNet model: in visually ambiguous or low-contrast environments, LiDAR and RADAR provide robust spatial and reflective information, whereas in well-lit scenes with clear visual cues, RGB becomes more dominant.

%%%%%%%%%%%%%%%%%%%%%%%%%%%%%%%%%%%%%%%%%%%%%%%%%%%%%%%%%%%%%%%%%%%%%%%%%%%%%%%%%%%%%%%%
\section{Discussion \& Limitations}
 
This work introduced LRC-WeatherNet to address the challenge of real-time weather type classification using large-scale multimodal sensor data comprising LiDAR, RADAR, and camera inputs. 
Through extensive experiments on the MSU-4S dataset \cite{3}, we evaluated the performance of different unimodal and multimodal baseline approaches.
The results demonstrated that our middle fusion approach in LRC-WeatherNet, which combines early fused LiDAR and RADAR data streams with camera input, consistently achieved the highest classification accuracy with real-time performance. %This approach proved especially effective in distinguishing weather conditions with high visual and structural similarity, notably within the same season.
Further analysis of LRC-WeatherNet revealed its ability to dynamically adjust attention across modalities based on scene context. Camera input was dominant in visually clear conditions, whereas fused LiDAR and RADAR data became more influential in ambiguous scenarios. %However, inclusion of a RADAR-only stream occasionally led to redundancy and degraded performance for specific classes.

Our LRC-WeatherNet framework employs a uniform architecture across modalities, utilizing separate yet structurally identical EfficientNet-B0 branches for RGB and fused LiDAR-RADAR inputs. While this leads to a higher combined parameter count of 19.17 million, the compute demands are far significantly lower at 1.26 GMACs, with a much faster inference time of 8.27 milliseconds.

From a performance standpoint, LRC-WeatherNet-PP achieves slightly higher test accuracy (87.77\% vs. 86.66\%) and macro-averaged F1 score (0.87 vs. 0.85), likely due to its superior ability to model fine-grained 3D structure from LiDAR and RADAR data. However, LRC-WeatherNet delivers strong generalization with over nine times faster inference and significantly lower computational cost. %The differences in modality-specific performance also reflect architectural alignment: LiDAR-only classification is stronger in PointPillars (81.49\%) than in EfficientNet-B0 (56.34\%), while Camera-only classification favors EfficientNet-B0 (77.86\%) over the EfficientNet-B7 branch used in the PointPillars pipeline (65.74\%). This indicates that EfficientNet-B0 is more effective at extracting semantic features from 2D images, while PointPillars is better optimized for geometric reasoning.

%The design of the Gated Fusion modules in each framework further reflects these architectural and input differences. In the LRC-WeatherNet pipeline, fusion occurs after global average pooling, where each modality's feature map is condensed into a flattened vector of shape $[B, C]$. The Gated Fusion module here uses \texttt{nn.Linear} layers to compute modality-specific gate values and fuses the RGB and fused LiDAR-RADAR features at the vector level. This design is lightweight and well-suited to late-stage fusion, where high-level semantic representations dominate.

%In contrast, the PointPillars pipeline performs fusion earlier in the network, directly on 2D spatial feature maps of shape $[B, C,- \\H, W]$. To preserve spatial granularity, the corresponding Gated Fusion module uses \texttt{nn.Conv2d} layers, allowing it to compute gating weights across both channel and spatial dimensions. This spatially-aware gating mechanism ensures that fusion is localized, enabling the model to dynamically emphasize or suppress modality contributions at specific spatial regions. These design choices were made to align with each framework's internal architecture: vector-level gating for EfficientNet's globally pooled features, and spatial-level gating for PointPillars' convolutional backbones.

Although the LRC-WeatherNet-PP baseline achieves higher overall accuracy across most classes, it exhibits increased confusion among fall-related classes and signs of overfitting in those categories. In contrast, our LRC-WeatherNet model demonstrates more balanced class-wise performance. The confusion matrix in Fig.~\ref{fig:confmat_bestfusion} indicates improved discriminability of LRC-WeatherNet among visually and structurally similar classes. While its overall accuracy is slightly lower than that of the PointPillars-based fusion model, LRC-WeatherNet offers enhanced generalization and robustness in distinguishing fall-related seasonal categories.

Overall, this work confirms the value of multimodal fusion for robust and accurate weather classification. The findings emphasize the importance of selecting appropriate fusion strategies and model architectures, underscoring the benefit of context-sensitive sensor integration in enhancing performance under diverse environmental conditions.

\textbf{Limitation:}
The selection of the dataset was primarily influenced by the lack of publicly available options that provide synchronized LiDAR, RADAR, and RGB camera data. 
Specifically, the MSU-4S dataset \cite{3} includes only one class representing rain and two classes corresponding to snowy conditions. The remaining categories depict seasonal environments, such as spring, summer, and fall, but without active precipitation. These seasonal scenes are defined more by ambient features, such as foliage color and lighting, rather than by dynamic weather phenomena like fog, heavy rainfall, hail, or snowstorms. As a result, the model has limited exposure to the types of weather conditions that most impact sensor reliability and visibility. Additionally, the dataset lacks rare yet critical edge cases, such as dense fog, sleet, hail, or nighttime snow, which are essential for training and evaluating perception systems intended for real-world deployment.

%\bibliographystyle{IEEEtran}
%\bibliography{IEEEabrv,bib}

%%%%%%%%%%%%%%%%%%%%%%%%%%%%%%%%%%%%%%%%%%%%%%%%%%%%%%%%%%%%%%%%%%%%%%%%%%%%%%%%

\end{document}